\documentclass[preprint,12pt]{elsarticle}

\usepackage{newtxtext}
\usepackage[normalem]{ulem}
\useunder{\uline}{\ul}{}

\journal{Applied Intelligence}

\usepackage{geometry}
\usepackage{url}
\usepackage{ upgreek }
\usepackage{pdflscape}
\usepackage{caption}
\usepackage{setspace}
\usepackage{pifont}
\usepackage{amsmath}
\usepackage{bbding}
\usepackage{amsfonts} 
\usepackage{amssymb}
\usepackage{mathrsfs}
\usepackage{lineno}
\usepackage{color}
\usepackage{ dsfont }
\usepackage[misc]{ifsym}

\usepackage{algorithm}
\usepackage{algorithmicx}
\usepackage{algpseudocode}
\usepackage{amsmath}      
\usepackage{graphicx}  
\usepackage{array}
\usepackage{verbatim}
\usepackage{titletoc}
\usepackage{color,xcolor}

\usepackage{units}
\usepackage{multirow}
\usepackage{tabularx}
\usepackage{booktabs}
\usepackage{bbm}
\usepackage{amsmath}      
\usepackage{bm} 
\usepackage{hyperref}
\usepackage[caption=false,font=scriptsize,labelfont=sf,textfont=sf]{subfig}

\usepackage{amsmath}

\usepackage{xcolor}
\definecolor{DeBlue}{RGB}{68,114,196}
\definecolor{DeGreen}{RGB}{84,130,53}
\definecolor{DeBlack}{RGB}{0,0,0}

\usepackage{natbib}
\usepackage{fancyhdr}

\usepackage{amsmath}
\allowdisplaybreaks[2]

\begin{document}

\begin{frontmatter}

\title{An Attention-Based Algorithm for Gravity Adaptation Zone Calibration}

\author[inst1]{Chen Yu\corref{cor1}}

\affiliation[inst1]{organization={Northeast Normal University},
            addressline={Northeast Normal University}, 
            city={Changchun, Jilin},
            country={China}}


         
\cortext[cor1]{Chen Yu is the corresponding author (email: yuc@nenu.edu.cn)}

\normalsize
\begin{abstract}
Accurate calibration of gravity adaptation zones is of great significance in fields such as underwater navigation, geophysical exploration, and marine engineering. With the increasing application of gravity field data in these areas, traditional calibration methods based on single features are becoming inadequate for capturing the complex characteristics of gravity fields and addressing the intricate interrelationships among multidimensional data. This paper proposes an attention-enhanced algorithm for gravity adaptation zone calibration. By introducing an attention mechanism, the algorithm adaptively fuses multidimensional gravity field features and dynamically assigns feature weights, effectively solving the problems of multicollinearity and redundancy inherent in traditional feature selection methods, significantly improving calibration accuracy and robustness.In addition, a large-scale gravity field dataset with over 10,000 sampling points was constructed, and Kriging interpolation was used to enhance the spatial resolution of the data, providing a reliable data foundation for model training and evaluation. We conducted both qualitative and quantitative experiments on several classical machine learning models (such as SVM, GBDT, and RF), and the results demonstrate that the proposed algorithm significantly improves performance across these models, outperforming other traditional feature selection methods. The method proposed in this paper provides a new solution for gravity adaptation zone calibration, showing strong generalization ability and potential for application in complex environments. The code is available at \href{this link}
{https://github.com/hulnifox/RF-ATTN}.
\end{abstract}


\end{frontmatter}


\section{Introduction}
Accurate calibration of gravity adaptation zones is of great significance in fields such as underwater navigation, geophysical exploration, and marine engineering \citep{zimmerman1999experimental}. In complex underwater environments, accurate gravity adaptation zone calibration not only improves the positioning accuracy of navigation systems but also effectively ensures the safety and efficiency of underwater operations \citep{zimmerman1999experimental,zago2005internal}. With the continuous advancement of underwater technology, traditional navigation methods can no longer meet modern requirements for high precision and reliability. Therefore, researchers have gradually realized the potential of gravity field data in underwater navigation, which can effectively assist in positioning and path planning by providing information related to variations in the Earth's gravity field \citep{zago2005fast}. Moreover, accurate gravity adaptation zone calibration can help identify suitable operational areas, optimize resource allocation, and reduce time and costs in operations. Therefore, the development of more precise calibration methods is not only of great value for scientific research but also plays a crucial role in technological advancements and economic benefits in practical applications \citep{yakowitz1985comparison}.

In recent years, research on gravity adaptation zone calibration has gradually increased, with various methods adopted to improve calibration accuracy. Early studies mainly focused on threshold and information entropy methods based on single feature indicators. Although these methods could assess adaptation to some extent, they often failed to capture the complex characteristics of the gravity field \citep{nabighian2005historical}. As the research progressed, Xiao et al. (2001) proposed multidimensional gravity field feature parameters, including gravity field standard deviation, roughness, and correlation coefficients, providing richer information for gravity adaptation zone calibration. Later, Cheng Li et al. (2007) and Cai Tijun et al. (2013) further developed selection criteria for gravity adaptation zones based on statistical features and analytic hierarchy processes, significantly improving the reasonableness and effectiveness of calibration \citep{crossley2013measurement}.

Meanwhile, the Earth Gravitational Model (EGM) serves as a high-precision gravity field model, providing strong support for global gravity field research and applications \citep{gundogdu2007spatial}. EGM2008 is currently the most widely used version, covering global gravity field data and corrected through various means, including satellite and ground-based gravity measurements \citep{nowell1999gravity}. The accuracy and reliability of the EGM model make it an important reference for gravity adaptation zone calibration \citep{he2024research}. With the continuous improvement in data collection technology, future EGM models are expected to achieve higher spatial resolution and accuracy, providing a solid data foundation for the application of underwater navigation and geographic information systems \citep{sobrero2024robust}. Overall, research on gravity adaptation zone calibration is moving toward more precise and comprehensive approaches \citep{bonvalot1998continuous}, with the EGM model playing an indispensable role in this process.

Despite progress in gravity adaptation zone calibration research, several issues remain, including a lack of high-quality, high-resolution data. Although the EGM model provides global gravity field data, data scarcity and uneven distribution in certain specific regions, especially in underwater environments or complex terrains, can lead to inaccuracies in calibration results \citep{kirby1974theoretical,yakowitz1985comparison}. This data deficiency limits the model's generalization ability in specific regions, affecting the accurate identification of adaptation zones and navigation performance. Moreover, existing calibration methods often struggle to dynamically adjust relationships between features, lacking sufficient generalization. Traditional feature selection and fusion methods usually rely on static feature indicators, failing to effectively handle complex interactions and nonlinear relationships between features \citep{nowell1999gravity,nabighian2005historical}. This limits the model's adaptability in changing environments, particularly in cases where gravity field characteristics are highly variable, leading to reduced calibration accuracy \citep{bonvalot1998continuous,oliver1990kriging}. Therefore, there is an urgent need to develop more flexible feature fusion methods that can adaptively adjust relationships between features, improving the model's performance and generalization ability in different environments.

Our model, based on an attention-enhanced feature fusion method, aims to improve the accuracy and robustness of gravity adaptation zone calibration. The model dynamically adjusts feature weights to effectively capture the complex interrelationships between gravity field features. Specifically, the model first processes the input multidimensional features, including gravity field standard deviation, roughness, and correlation coefficients, and then introduces an attention mechanism to flexibly adjust the influence of each feature in a specific environment. This method not only overcomes the common multicollinearity problem in traditional feature selection but also significantly improves the ability to identify key features.

In summary, the main contributions of this paper are as follows:
\begin{itemize}
    \item We propose a new attention-enhanced algorithm for gravity adaptation zone calibration that adaptively fuses multiple features. By introducing an attention mechanism, the algorithm dynamically assigns feature weights to better capture the complex relationships between features. This method effectively overcomes the multicollinearity and redundancy issues common in traditional feature selection methods, significantly improving the accuracy and robustness of adaptation zone calibration.
    \item We constructed a large-scale dataset specifically for gravity adaptation zone calibration, covering four different areas with over 10,000 real sampling points. The dataset, processed through Kriging interpolation, has high spatial resolution, providing rich and realistic gravity field data for model training and evaluation. This dataset provides important data support for gravity adaptation zone calibration research, filling a gap in the field regarding high-quality datasets.
    \item We conducted comprehensive qualitative and quantitative experiments. The experimental results show that the proposed algorithm consistently improves performance across multiple classical machine learning models and significantly outperforms other feature selection algorithms. Whether in traditional models (such as SVM, GBDT) or ensemble models (such as RF, ADABOOST), the introduction of attention-enhanced feature fusion methods leads to varying degrees of performance improvement, validating the wide applicability and strong feature representation capabilities of the algorithm.
\end{itemize}
\section{Related Work}
\subsection{Earth Gravity Field Models}
The development of pre-trained models has significantly improved the performance of gravity adaptation zone calibration tasks. Early gravity field models such as EGM2008 \citep{he2024research} provided a high-resolution reference model of Earth's gravity field. EGM2008 utilized satellite, ground gravity, and topographic data from around the globe, extending the degree and order of spherical harmonic expansion to 2190, making it a high-resolution global static gravity field benchmark model. This model is widely used in geographic information systems and various gravity field applications, laying the foundation for subsequent research. EGM2020, an improved version of EGM2008, is expected to include updated global data with higher accuracy. Similar to EGM2008, EGM2020 will continue to use spherical harmonic expansion, but its improvements lie in the integration of new data from satellite missions such as GOCE and GRACE, further enhancing the model’s spatial resolution and gravity field accuracy. The improved accuracy of EGM2020, especially in high-resolution global gravity field modeling, significantly enhances the capability of gravity field feature analysis, making it more suitable for precise gravity adaptation zone calibration. Based on these pre-trained gravity models, researchers have gradually explored ways to further improve the accuracy of gravity field adaptation analysis in gravity matching navigation tasks. Li et al. (2020) proposed a gravity matching navigation adaptability analysis method based on EGM2008, generating a gravity adaptability distribution map through interpolation and densification of marine gravity field characteristic parameters, thereby automatically calibrating adaptation zones. Wang et al. (2021) proposed an improved adaptability analysis method based on EGM2020, utilizing the high resolution and multi-scale analysis capabilities of EGM2020 to effectively improve the accuracy and robustness of adaptation zone identification. The continuous updates of these pre-trained models indicate that as data precision and model complexity improve, the adaptability analysis of gravity field products will continue to be optimized. However, analysis methods based on static gravity field features still have limitations, especially when dealing with dynamic observation data.

\subsection{Gravity Adaptation Zone Calibration Methods}
The development of gravity adaptation zone calibration can be traced back to early threshold and information entropy methods based on single features \citep{kirby1974theoretical}. These methods measured the adaptability of underwater gravity fields through simple feature analysis but were limited in handling complex gravity fields. In 2001, Xiao et al. proposed a multi-dimensional gravity field feature parameter analysis method, including gravity field standard deviation, roughness, correlation coefficient, slope standard deviation, skewness coefficient, and gravity anomaly difference entropy. These parameters provided a more comprehensive reflection of the complexity and variability of the gravity field, marking a significant advancement in gravity adaptation zone calibration. Subsequently, Cheng Li et al. (2007) proposed a gravity adaptation zone selection method based on local statistical features, while Cai Tijun et al. (2013) introduced an analytic hierarchy process for a comprehensive evaluation of gravity matching regions, further improving the rationality of adaptation zone calibration. Ouyang et al. (2020) used multi-attribute decision theory to weight gravity field features, addressing the problem of feature weight allocation. However, the issue of overlap and collinearity among multi-attribute features still exists. Zhang et al. (2021) introduced an improved A* algorithm to optimize route planning and non-adaptive zone avoidance, further improving navigation accuracy \citep{nabighian2005historical}. In recent years, with the rise of deep learning and attention mechanisms, attention-based feature fusion methods have brought new breakthroughs in gravity adaptation zone calibration. These methods can adaptively adjust feature weights, effectively handling complex feature relationships, improving the accuracy and robustness of calibration, and becoming a key direction for future gravity adaptation zone calibration.

\subsection{Preliminary}
\label{p}
In order to better describe the characteristics of the gravity field and the gradient field in the gravity adaptation zone, the following statistical indicators are introduced in this paper:

\textbf{The standard deviation of the gravity field }($\sigma$)is used to measure the degree of dispersion of the gravity anomalies. The calculation formula is as follows:
\[
\sigma = \sqrt{\frac{1}{mn-1} \sum_{i=1}^{m}\sum_{j=1}^{n} \left[ \Delta g(i,j) - \overline{\Delta g} \right]^2}
\]
Where $\Delta g(i,j)$ represents the gravity anomaly at point $(i,j)$, and $\overline{\Delta g}$ is the mean of the gravity anomalies.

\textbf{Gravity Field Roughness} ($r$) is used to describe the degree of fluctuation in the gravity field along the longitude and latitude directions. It is defined as the average of the roughness in the longitude direction $r_{\lambda}$ and the roughness in the latitude direction $r_{\phi}$:
\[
r = \frac{r_{\lambda} + r_{\phi}}{2}
\]

\textbf{Gravity Field Correlation Coefficient} ($R$) is used to evaluate the correlation of the gravity field in the longitude and latitude directions, and the formula is as follows:
\[
R = \frac{R_{\lambda} + R_{\phi}}{2}
\]

\textbf{Gradient Standard Deviation}: For the analysis of the gradient field, the gradient standard deviation ($\sigma_S$) is used to measure the variation in the gradient field. The calculation formula is:
\[
\sigma_S = \sqrt{\frac{1}{mn-1} \sum_{i=1}^{m}\sum_{j=1}^{n} \left[ S(i,j) - \overline{S} \right]^2}
\]
where $S(i,j)$ represents the gradient field value, and $\overline{S}$ is the mean value of the gradient field.

\textbf{Skewness Coefficient}: To analyze the symmetry of the gravity field, the skewness coefficient ($C_s$) is used, and the formula is:
\[
C_s = \frac{mn}{(m-1)(m-2)(n-1)(n-2)} \cdot \frac{1}{\sigma^3} \sum_{i=1}^{m} \sum_{j=1}^{n} \left( \Delta g(i,j) - \overline{\Delta g} \right)^3
\]
This coefficient reflects the degree of skewness in the distribution of gravity anomalies.

\textbf{Gravity Anomaly Differential Entropy} ($H$) is used to measure the complexity of the gravity field, and the formula is as follows:
\[
H = -\sum_{i=1}^{m}\sum_{j=1}^{n} P_{ij} \log_2 P_{ij}
\]
where $P_{ij}$ is the probability distribution at point $(i,j)$.

\subsection{Model Overview}
\begin{figure}[htbp]
    \centering
    \includegraphics[width=1\linewidth]{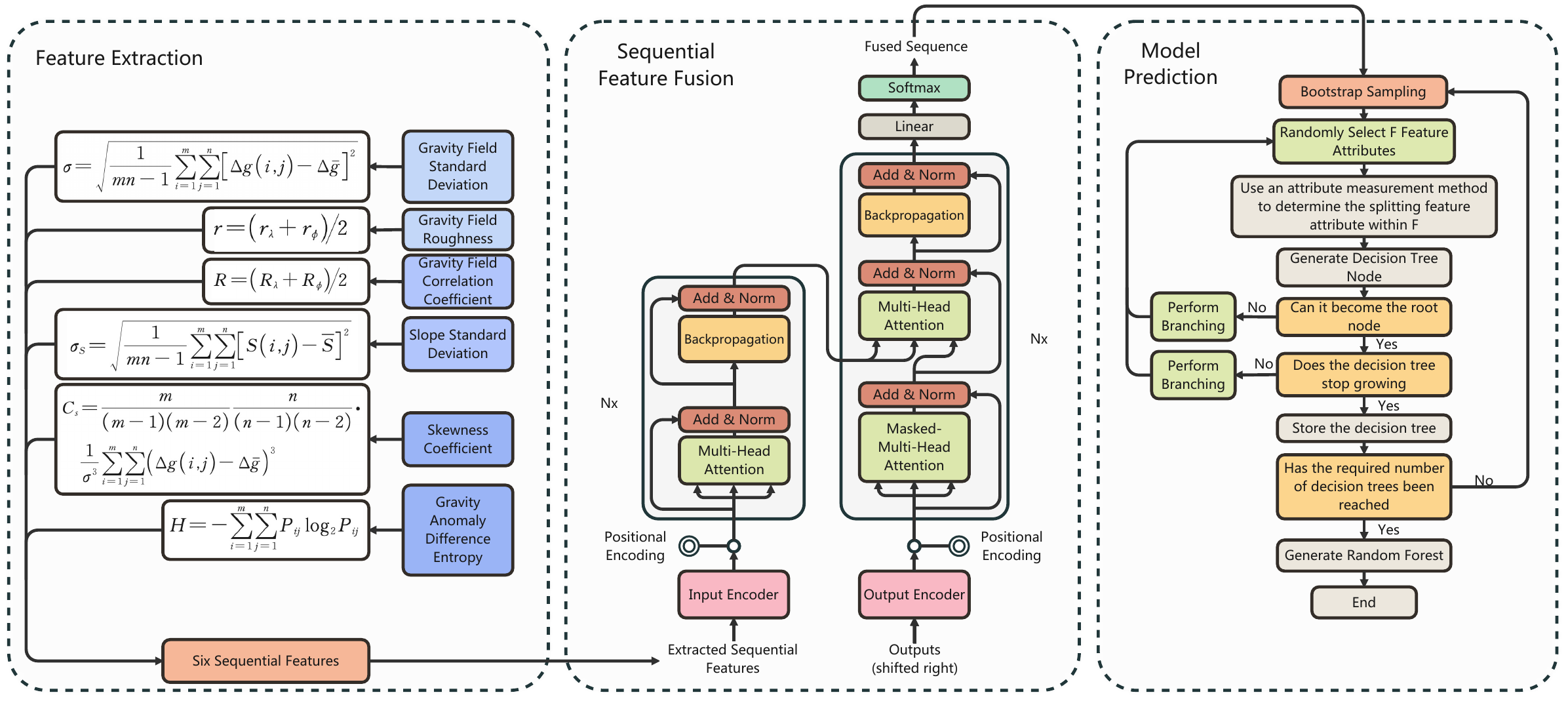}
    \caption{Overview of our proposed model, \textbf{Left}: Feature extraction from gravity anomaly data using the method described in Section \ref{p}, \textbf{Middle}: Feature fusion method based on the attention mechanism, \textbf{Right}: Calibration of the gravity adaptation zone using the Random Forest model with the extracted features.}
    \label{fig:zl}
\end{figure}
The overall framework of our proposed model is shown in Figure \ref{fig:zl}. The model extracts various features from gravity anomaly data, including statistical features, spatial distribution features, and more.  
In the feature fusion stage, we designed an attention mechanism-based strategy that allows the model to adaptively focus on important feature dimensions, improving its ability to interpret complex gravity field data. This approach enables the model to effectively reduce the impact of noise on calibration and enhance the significance of key features in the calibration process. The fused features are then input into the Random Forest model for gravity adaptation zone calibration.

\subsection{Feature Fusion}
As shown in Figure \ref{fig:zl}, the feature fusion stage of our model is based on an attention mechanism. Given the input feature set $\mathbf{X} = \{ \mathbf{x}_1, \mathbf{x}_2, \dots, \mathbf{x}_n \}$, where $\mathbf{x}_i \in \mathbb{R}^d$ represents the $i$-th feature vector, $n$ is the total number of features, and $d$ is the dimension of each feature. Each $\mathbf{x}_i$ contains multiple features extracted from gravity anomaly data, including the gravity field's standard deviation, roughness, correlation coefficient, gradient standard deviation, skewness coefficient, and entropy.

To adaptively adjust the importance of each feature, we assign a weight to each feature using an attention mechanism. First, we define the weight coefficient as $\alpha_i$, representing the weight of each feature. The calculation of the weight coefficient depends on the similarity between features, which is evaluated through a scoring function that assesses the relationships between features. Specifically, we use a dot-product form to calculate the similarity score between feature $\mathbf{x}_i$ and feature $\mathbf{x}_j$:
\[
e_{ij} = \mathbf{x}_i^T \mathbf{W} \mathbf{x}_j
\]
where $\mathbf{W} \in \mathbb{R}^{d \times d}$ is a learnable weight matrix, and $e_{ij}$ represents the similarity score between feature $\mathbf{x}_i$ and feature $\mathbf{x}_j$.

To ensure the importance of different features is normalized, we apply a Softmax function to the similarity scores $e_{ij}$ to compute the attention weight $\alpha_i$ for each feature:
\[
\alpha_i = \frac{\exp(e_{ij})}{\sum_{j=1}^{n} \exp(e_{ij})}
\]
where $\alpha_i$ represents the normalized weight of the $i$-th feature, ensuring that the sum of all weights equals 1, i.e., $\sum_{i=1}^{n} \alpha_i = 1$.

Finally, the feature fusion is completed through a weighted summation, resulting in the fused feature vector $\mathbf{z}$, defined as:
\[
\mathbf{z} = \sum_{i=1}^{n} \alpha_i \mathbf{x}_i
\]
where $\mathbf{z}$ is the final fused feature vector, which integrates the weight information of different features. This fused feature will be used in the subsequent gravity adaptation zone calibration model, where prediction and calibration are performed using a Random Forest classifier.

\subsection{Feature Prediction}

For the feature prediction stage, we selected a Random Forest classifier as the final classification model. The Random Forest classifier, which trains by constructing multiple decision trees, is effective in handling high-dimensional feature data. After training, we used this classifier to predict the fused features and determine the adaptability within the study area. To identify highly adaptive navigation areas, we adopted the lower quantile of gravity anomaly values as the classification criterion. Specifically, we calculated the lower quantile of the gravity anomaly values and defined points with quantile values below 0.05 as high-adaptability areas. Using this method, we identified regions with low gravity anomalies that are suitable for navigation.

\subsection{Training Process}

After feature extraction and fusion, the training process is outlined in the following pseudo-code section. Specifically, during the model training process, we divided the fused features into training and testing sets and standardized the features to ensure that they are on the same scale, preventing the differences in feature values from affecting the model's training performance. During the model training stage, we used a Random Forest classifier, constructing multiple decision trees to enhance the model's generalization capability. In this process, we performed cross-validation to tune the model's hyperparameters, including the number of trees and the maximum depth, to improve the model's accuracy and prevent overfitting.

In the model evaluation stage, we used the test set to make predictions and evaluated the model's performance through classification reports and confusion matrices, which demonstrated the model's accuracy and robustness in calibrating gravity adaptation zones. Finally, through feature importance analysis, we identified the features that contributed most to the classification results and visualized the model's prediction results, showing its application in gravity adaptation zone calibration.

\begin{algorithm}[H]
\caption{Model Training and Feature Fusion Process}
\begin{algorithmic}[1]
    \State \textbf{Input}: Gravity anomaly data $\mathbf{X} = \{ \mathbf{x}_1, \mathbf{x}_2, \dots, \mathbf{x}_n \}$, where $\mathbf{x}_i \in \mathbb{R}^d$ is the feature vector, $n$ is the total number of features, and $d$ is the feature dimension.
    \State \textbf{Output}: Optimized Random Forest model $clf$ for gravity adaptation zone calibration.

    \State \textbf{Step 1: Feature Extraction}
    \For{each grid point $i$}
        \State Calculate gravity field standard deviation $\sigma$
        \State Calculate roughness $r$
        \State Calculate correlation coefficient $R$
        \State Calculate gradient standard deviation $\sigma_S$
        \State Calculate skewness coefficient $C_s$
        \State Calculate entropy $H$
    \EndFor

    \State \textbf{Step 2: Feature Fusion Based on Attention Mechanism}
    \For{each feature vector $\mathbf{x}_i$}
        \State Compute the feature similarity score $e_{ij}$:
        \Statex $e_{ij} = \mathbf{x}_i^T \mathbf{W} \mathbf{x}_j$
        \State Compute the attention weight $\alpha_i$ using Softmax:
        \Statex $\alpha_i = \frac{\exp(e_{ij})}{\sum_{j=1}^{n} \exp(e_{ij})}$
        \State Fuse features based on the weights $\alpha_i$:
        \Statex $\mathbf{z} = \sum_{i=1}^{n} \alpha_i \mathbf{x}_i$
    \EndFor

    \State \textbf{Step 3: Random Forest Model Training and Optimization}
    \State Split the dataset into training and testing sets, and perform feature scaling.
    \State Train the model using a Random Forest classifier:
    \Statex $clf = \text{RandomForestClassifier}(n\_estimators=100, random\_state=42)$
    \State Use cross-validation to select the best hyperparameters and train the optimized model.

    \State \textbf{Step 4: Model Evaluation and Prediction}
    \State Make predictions on the test set and compute the classification report and confusion matrix.
    \State Analyze the feature importance from the Random Forest model to determine the most important features.
\end{algorithmic}
\end{algorithm}

\section{Experiments}
\subsection{Dataset Construction}
We evaluated the model using four regions from the South China Sea. The geographic ranges of the datasets are as follows:
\begin{itemize}
    \item Region 1: 19.48°N - 22.00°N, 118.38°E - 120.60°E
    \item Region 2: 14.18°N - 16.10°N, 116.93°E - 119.00°E
    \item Region 3: 12.30°N - 16.10°N, 109.40°E - 122.20°E
    \item Region 4: 10.00°N - 13.40°N, 109.40°E - 122.20°E
\end{itemize}
Gravity anomaly data for all regions were generated using the EGM2008 model, and relevant indicators were calculated according to the formulas in Section \ref{p}. The 3D topography maps of the four regions are shown below:
\begin{figure}[H]
    \centering
    \includegraphics[width=0.8\linewidth]{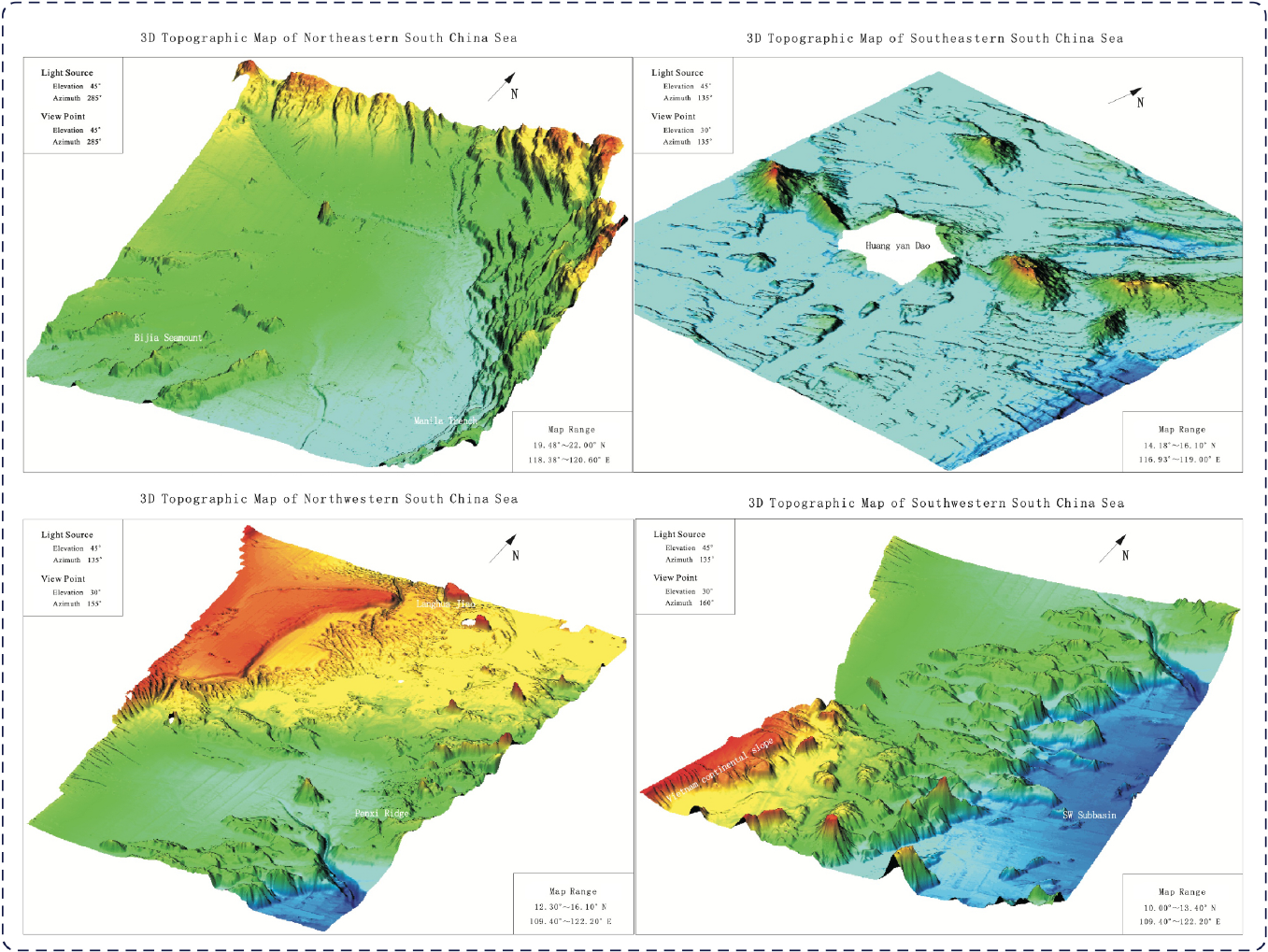}
    \caption{Topography maps of the four selected sea areas}
    \label{fig:dx}
\end{figure}
For the obtained data, comprehensive data preprocessing was first conducted. During the preprocessing phase, we performed cleaning and normalization of the gravity anomaly data from the sampling regions. Since geographic data often contains a certain degree of noise and missing values, we ensured data quality and consistency through interpolation and outlier handling techniques. Specifically, we applied Kriging Interpolation to the gravity anomaly data collected from each sampling region. Kriging Interpolation is capable of predicting unobserved points based on known sampling point data, making it particularly suitable for dealing with irregular distributions in geospatial data.

This method is based on the spatial autocorrelation between sample points, assuming that geographic phenomena change smoothly. It estimates the values of unknown points by taking a weighted average of the known points, with the weights determined by the distance between sampling points and spatial variability. In the specific implementation, we calculated the semivariogram for the gravity anomaly sampling points in each region. The semivariogram describes the spatial variability, with similarity between sampling points decreasing as distance increases. Based on this variability model, we performed Kriging Interpolation to predict gravity anomaly values for unobserved points in each sampling region.

Before and after interpolation, we compared the smoothness and spatial continuity of the data. The data before interpolation was limited to a finite number of sampling points and exhibited some degree of spatial discontinuity. However, after Kriging Interpolation, the data exhibited a smoother spatial variation trend, and gravity anomaly values were obtained for each grid point in all regions.

\begin{figure}[H]
    \centering
    \includegraphics[width=1\linewidth]{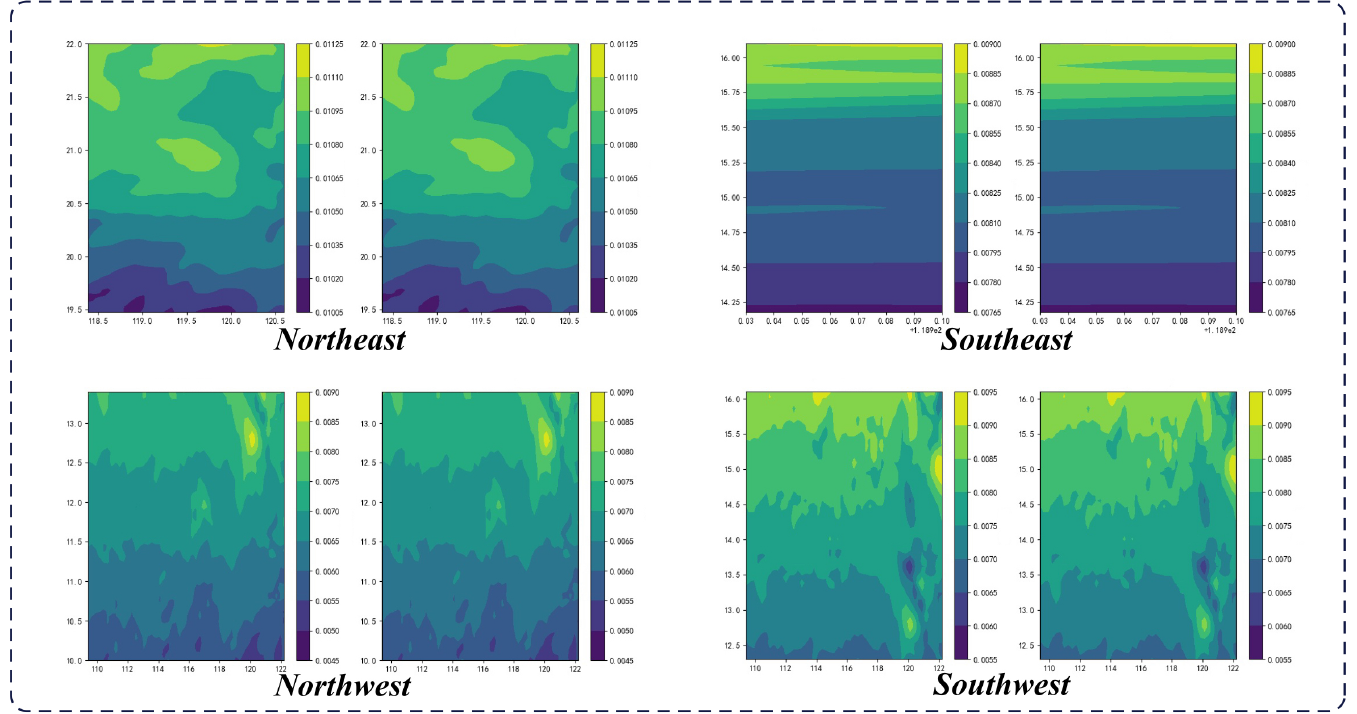}
    \caption{Heatmap of gravity anomaly values obtained from the EGM2008 model for the four selected sea regions, \textbf{Left}: Before interpolation, \textbf{Right}: After interpolation}
    \label{fig:dx}
\end{figure}
\subsection{Evaluation Metrics}

To comprehensively evaluate the classification performance of our model in the gravity adaptation zone calibration task, we selected the following three commonly used evaluation metrics: Accuracy, Recall, and F1 Score. These metrics provide different perspectives to assess the model's performance, ensuring good classification capability for both positive classes (adaptation zones) and negative classes (non-adaptation zones).

\subsubsection{Accuracy}
Accuracy is used to measure the proportion of correctly classified samples among all test samples. The calculation formula is as follows:
\[
\text{Accuracy} = \frac{TP + TN}{TP + TN + FP + FN}
\]
where:
\begin{itemize}
    \item $TP$ (True Positive): The number of samples correctly predicted as positive, i.e., the number of true adaptation zone samples correctly classified as adaptation zones.
    \item $TN$ (True Negative): The number of samples correctly predicted as negative, i.e., the number of true non-adaptation zone samples correctly classified as non-adaptation zones.
    \item $FP$ (False Positive): The number of samples incorrectly predicted as positive, i.e., the number of true non-adaptation zone samples incorrectly classified as adaptation zones.
    \item $FN$ (False Negative): The number of samples incorrectly predicted as negative, i.e., the number of true adaptation zone samples incorrectly classified as non-adaptation zones.
\end{itemize}

\subsubsection{Recall}
Recall is an indicator used to measure the model's ability to recognize positive classes (adaptation zones). It represents the proportion of true positive samples that are correctly predicted among all actual positive samples. The calculation formula is:
\[
\text{Recall} = \frac{TP}{TP + FN}
\]
\subsubsection{F1 Score}
F1 Score is the harmonic mean of Precision and Recall, used to comprehensively evaluate the classification performance of the model, particularly effective in cases of imbalanced data. The calculation formula is:
\[
F1 = 2 \cdot \frac{\text{Precision} \cdot \text{Recall}}{\text{Precision} + \text{Recall}}
\]
where Precision is defined as:
\[
\text{Precision} = \frac{TP}{TP + FP}
\]
\subsection{Comparison Models}
In this section, we selected comparison models to answer two primary questions:
\begin{itemize}
    \item \textbf{Q1}: How well does the attention-enhanced feature fusion method apply to different algorithms?
    \item \textbf{Q2}: Are there performance differences between different feature selection methods and the attention-based feature fusion method?
\end{itemize}

\textbf{RQ1:} We selected four classic machine learning models to validate our method:
\begin{itemize}
    \item \textbf{SVM (Support Vector Machine)}: SVM is a widely used classification method that separates data into different classes by finding an optimal hyperplane.
    \item \textbf{GBDT (Gradient Boosting Decision Tree)}: GBDT builds multiple weak classifiers progressively and optimizes errors at each step to improve the overall classification performance of the model.
    \item \textbf{RF (Random Forest)}: Random Forest builds multiple decision trees and determines the final classification result through majority voting. It is robust against noise and has strong generalization capabilities.
    \item \textbf{ADABOOST}: The Adaptive Boosting algorithm iteratively adjusts the weights of weak classifiers to gradually improve the model's ability to recognize hard-to-classify samples.
\end{itemize}

\textbf{RQ2:} We designed several variants based on RF, primarily using different feature selection methods, including:
\begin{itemize}
    \item \textbf{Pearson Feature Selection}: This method uses the Pearson correlation coefficient to select features that have high linear correlation with the target variable.
    \item \textbf{SHAP Feature Selection}: SHAP (Shapley Additive Explanations) values are used to measure each feature's contribution to the model's output, providing a ranking for feature importance and selection.
    \item \textbf{LASSO Feature Selection}: This method uses L1 regularization to sparsely select features, effectively reducing the number of features and enhancing model interpretability.
    \item \textbf{Relief Feature Selection}: Features are scored based on their ability to distinguish between neighboring samples, selecting those with higher discriminative power.
    \item \textbf{RF-ATTN}: A feature fusion method based on Random Forest and attention mechanism, where weights are adaptively assigned to different features, improving the model's classification performance on complex data.
\end{itemize}

\subsection{Experimental Results}
\begin{table}[H]
\small
\centering
\caption{Performance of the attention-enhanced gravity adaptation zone calibration algorithm on various algorithms, with bold indicating the best performance.}
\resizebox{\textwidth}{!}{
\begin{tabular}{c!{\vrule}ccc!{\vrule}ccc!{\vrule}ccc!{\vrule}ccc}
\toprule
\label{table1}

\textbf{Model}     & \multicolumn{3}{c!{\vrule}}{\textbf{Northeast}}   & \multicolumn{3}{c!{\vrule}}{\textbf{Southeast}}   & \multicolumn{3}{c!{\vrule}}{\textbf{Northwest}}   & \multicolumn{3}{c}{\textbf{Southwest}}             \\ 
\cmidrule(l{2pt}r{2pt}){2-4} \cmidrule(l{2pt}r{2pt}){5-7} \cmidrule(l{2pt}r{2pt}){8-10} \cmidrule(l{2pt}r{2pt}){11-13}
                     & \textbf{Acc}  & \textbf{F1} & \textbf{Recall}   & \textbf{Acc}  & \textbf{F1} & \textbf{Recall}   & \textbf{Acc}  & \textbf{F1} & \textbf{Recall}  & \textbf{Acc}  & \textbf{F1} & \textbf{Recall}
                     \\ 
\midrule
\textbf{SVM-ATTN}     & \textbf{66.9} & \textbf{80.2} & \textbf{66.9}   & \textbf{79.3} & \textbf{88.5} & \textbf{79.3}   & \textbf{79.2} & \textbf{87.7} & \textbf{79.2}   & \textbf{67.3} & \textbf{80.5} & \textbf{67.3}   \\
\textbf{SVM}          & 34.9          & 51.8          & 34.9            & 28.7          & 44.6          & 28.7            & 73.1          & 83.8          & 73.1            & 64.1          & 78.1          & 64.1            \\ \midrule
\textbf{GBDT-ATTN}    & \textbf{93.3} & \textbf{96.6} & \textbf{93.3}   & \textbf{90.5} & \textbf{95.0} & \textbf{90.5}   & \textbf{84.4} & \textbf{90.8} & \textbf{84.4}   & \textbf{95.5} & \textbf{97.7} & \textbf{95.5}   \\
\textbf{GBDT}         & 93.2          & 96.0          & 93.2            & 89.9          & 94.7          & 89.9            & 84.1          & 90.7          & 84.1            & \textbf{95.5} & \textbf{97.7} & \textbf{95.5}   \\\midrule
\textbf{RF-ATTN}      & \textbf{95.7} & \textbf{97.8} & \textbf{95.7}   & \textbf{92.3} & \textbf{96.0} & \textbf{92.3}   & \textbf{90.9} & \textbf{94.5} & \textbf{90.9}   & \textbf{95.6} & \textbf{97.8} & \textbf{95.6}   \\
\textbf{RF}           & 94.3          & 97.0          & 94.3            & 92.0          & 95.8          & 92.0            & 89.1          & 98.5          & 93.5            & 95.5          & 97.7          & 95.5            \\\midrule
\textbf{ADABOST-ATTN} & \textbf{93.2} & \textbf{96.5} & \textbf{93.2}   & \textbf{49.1} & \textbf{65.8} & \textbf{49.1}   & \textbf{72.7} & \textbf{83.4} & \textbf{72.7}   & \textbf{93.9} & \textbf{96.8} & \textbf{93.9}   \\
\textbf{ADABOST}      & \textbf{93.2}          & \textbf{96.5}          & \textbf{93.2}            & 48.9          & 65.7          & 48.9            & \textbf{72.7} & \textbf{83.4} & \textbf{72.7}   & \textbf{93.9}          & \textbf{96.8}          & \textbf{93.9 }           \\ 

\bottomrule
\end{tabular}
}
\end{table}
Table \ref{table1} shows the classification results of different algorithms with and without the use of the attention-enhanced feature fusion method. From the results, we can observe that most algorithms experience a significant performance improvement when using the attention-enhanced feature fusion method. Specifically, for SVM, GBDT, and RF, the introduction of the attention mechanism clearly enhances accuracy, F1 score, and recall.

For example, after applying the attention mechanism, SVM's classification performance significantly improved across multiple regions, especially in the Southeast region, where accuracy increased from 28.7 to 79.3, and the F1 score rose from 44.6 to 88.5. This indicates that the attention mechanism effectively enhances SVM's feature representation capability. RF and GBDT also showed notable improvements, particularly GBDT in the Northeast region, where accuracy increased from 93.2 to 93.3, and the F1 score rose from 96.0 to 96.6. In contrast, the improvement for ADABOOST was smaller, but in specific regions like Southeast and Northwest, it still demonstrated some level of enhancement.

Overall, the effectiveness of the attention-enhanced feature fusion varies across different algorithms, but whether for simpler or more complex models, the performance generally improves after incorporating the attention mechanism. This suggests that the attention mechanism can effectively adaptively assign feature weights, increasing the model's focus on key features, thereby improving classification performance in the gravity adaptation zone calibration task.

\begin{table}[htbp]
\small
\centering
\caption{A comparison of different feature selection methods with attention-based feature selection, where bold indicates the best performance, and underline indicates the second-best performance.
}
\resizebox{\textwidth}{!}{
\begin{tabular}{c!{\vrule}ccc!{\vrule}ccc!{\vrule}ccc!{\vrule}ccc}
\toprule
\label{tabel2}
\textbf{Model}     & \multicolumn{3}{c!{\vrule}}{\textbf{Northeast}}   & \multicolumn{3}{c!{\vrule}}{\textbf{Southeast}}   & \multicolumn{3}{c!{\vrule}}{\textbf{Northwest}}   & \multicolumn{3}{c}{\textbf{Southwest}}             \\ 
\cmidrule(l{2pt}r{2pt}){2-4} \cmidrule(l{2pt}r{2pt}){5-7} \cmidrule(l{2pt}r{2pt}){8-10} \cmidrule(l{2pt}r{2pt}){11-13}
                     & \textbf{Acc}  & \textbf{F1} & \textbf{Recall}   & \textbf{Acc}  & \textbf{F1} & \textbf{Recall}   & \textbf{Acc}  & \textbf{F1} & \textbf{Recall}  & \textbf{Acc}  & \textbf{F1} & \textbf{Recall}
                     \\ 
\midrule
Pearson        & 92.1          & 94.7          & 94.3            & NA            & NA            & NA              & 78.9          & 86.0          & 88.5            & 89.5          & 93.0          & 93.8            \\
SHAP           & 93.9          & 95.9          & 95.5            & {\ul 86.7}    & {\ul 91.1}    & 91.0            & 89.2          & 92.9          & {\ul 97.4}      & 91.6          & 94.4          & 94.5            \\
Lasso          & 70.7          & 80.7          & 82.3            & 66.1          & 76.6          & 73.7            & 64.8          & 76.7          & 79.5            & 67.1          & 77.5          & 75.9            \\
ReliefF        & {\ul 95.2}    & {\ul 96.8}    & \textbf{97.0}   & 86.5          & 91.0          & {\ul 91.1}      & {\ul 90.5}    & {\ul 93.8}    & \textbf{97.8}   & {\ul 93.1}    & {\ul 95.4}    & \textbf{96.8}   \\
\textbf{Ours}  & \textbf{95.7} & \textbf{97.8} & {\ul 95.7}      & \textbf{92.3} & \textbf{96.0} & \textbf{92.3}   & \textbf{90.9} & \textbf{94.5} & 90.9            & \textbf{95.6} & \textbf{97.8} & {\ul 95.6}      \\
\bottomrule
\end{tabular}
}
\end{table}
Table \ref{tabel2} shows the comparison results between different feature selection methods and attention-enhanced feature selection methods in the gravity adaptation zone calibration task. It can be observed that the attention-enhanced feature selection method outperforms other methods in most regions, especially in the Southeast and Southwest regions, where it achieves the highest accuracy, F1 score, and recall rate. While ReliefF performs similarly to our model in some regions, the attention-enhanced feature selection method overall performs best, particularly in the Northeast region, where it achieves an accuracy of 95.7 and an F1 score of 97.8, demonstrating its strong performance in gravity adaptation zone calibration.

\subsection{Discussion} This paper proposes a gravity adaptation zone calibration method based on attention-enhanced feature fusion, and experimental results verify its effectiveness across various machine learning models. The research workflow includes dataset construction and preprocessing, feature extraction, attention-based feature fusion, model training and evaluation, and comparative experiments of different feature selection methods. In the data preprocessing stage, the Kriging interpolation method was used to process gravity anomaly data. Experimental results show that the attention mechanism improves classification performance in most models, particularly in SVM, GBDT, and RF, effectively addressing SVM's poor performance in high-dimensional data and enhancing the robustness of GBDT and RF. Compared to traditional feature selection methods (e.g., Pearson, SHAP, and LASSO), the attention-based feature fusion method performs better, capturing complex interactions between features. Additionally, the proposed method demonstrates strong robustness in addressing the data imbalance problem, improving the recognition ability of minority class samples by reasonably distributing weights.

The proposed attention-enhanced feature fusion method provides an efficient and accurate solution for gravity adaptation zone calibration, with significant advantages in handling high-dimensional and complex data. In the future, this method can be extended to other geographic data analysis tasks, and the attention mechanism can be further optimized to meet the requirements of different tasks and large-scale dataset applications.

\newpage
\bibliographystyle{elsarticle-num} 
\bibliography{main.bib}
\end{document}